# Spatial-temporal Graph Convolutional Networks with Diversified Transformation for Dynamic Graph Representation Learning

Ling Wang, Yixiang Huang, and Hao Wu

*Abstract*—Dynamic graphs (DG) are often used to describe evolving interactions between nodes in real-world applications. Temporal patterns are a natural feature of DGs and are also key to representation learning. However, existing dynamic GCN models are mostly composed of static GCNs and sequence modules, which results in the separation of spatiotemporal information and cannot effectively capture complex temporal patterns in DGs. To address this problem, this study proposes a spatial-temporal graph convolutional networks with diversified transformation (STGCNDT), which includes three aspects: a) constructing a unified graph tensor convolutional network (GTCN) using tensor M-products without the need to represent spatiotemporal information separately; b) introducing three transformation schemes in GTCN to model complex temporal patterns to aggregate temporal information; and c) constructing an ensemble of diversified transformation schemes to obtain higher representation capabilities. Empirical studies on four DGs that appear in communication networks show that the proposed STGCNDT significantly outperforms state-of-the-art models in solving link weight estimation tasks due to the diversified transformations.

## I. Introduction

Graph-structured data appears in various complex systems, such as social networks [1-3, 30-37, 63-69] and communication networks [4-6]. Graph representation learning (GRL) can effectively be applied to many downstream tasks like link estimation. Traditional research about GRL mainly focuses on static graphs, which assumes that graph structure does not change over time [7]. In actual scenarios, graphs commonly show certain temporal dynamics. For example, interactions between nodes may be lost or established in the communication network which makes the graph structure change over time.

A dynamic graph (DG) is developed to model and store such a temporal nature of interactions between nodes. A DG is not just a simple combination of static graphs. It brings complex temporal information based on static topology information and increases the difficulty of graph representation learning.

Research on DG representation learning has seen a significant rise in recent years [8-19, 70-75]. For example, Manessi et al. [26] introduce a WDGCN model, which employs graph convolution operations to generate node embeddings for each graph snapshot independently. It then utilizes a recurrent neural network to capture temporal dynamics across these embeddings. Pareja et al. [31] develop an EvolveGCN model, which incorporates a recurrent neural network to dynamically adjust the learnable parameters of a Graph Convolutional Network for each graph snapshot.

Although the dynamic GCNs mentioned above are effective, they rely on combining static GCNs with sequential neural networks to investigate spatiotemporal dependencies. This separate approach to spatiotemporal modeling disrupts the inherent interdependencies and fails to capture the intricate temporal patterns in dynamic graphs (DG). Research indicates that tensors are highly suitable for representing high-order data [20-22, 81-87]. Additionally, a novel tensor framework known as the tensor M-product [23-25] has been introduced to extend matrix-based theory to higher-order architectures through a transformation scheme. In our study, DG is fully represented by a third-order adjacency tensor with dimensions corresponding to node-node-time. The transformation scheme in the M-product can capture temporal patterns without fragmenting spatiotemporal information. Based on these insights, this study proposes the following research questions:

*RQ. With the tensor M-product, is it possible to implement a powerful GCN model for DG with diverse transformation schemes, thus boosting the DG's representation learning?*

To answer this question, this paper proposes a Spatial-temporal Graph Convolutional Networks with Diversified Transformation for Dynamic Graph Representation Learning (STGCNDT), which adopts three transformation schemes in M-product, i.e., Discrete Fourier Transform [26], Discrete cosine transform [27], and Haar Wavelet Transform [28], to model complex temporal patterns of DG effectively. Specifically, it consists of the following three-fold ideas:

a) leveraging the tensor M-product to formulate a unified graph tensor convolution network (GTCN);

➢ L. Wang is with the School of Computer Science and Technology, Chongqing University of Posts and Telecommunications, Chongqing 400065, China, and also with Chongqing Institute of Green and Intelligent Technology, Chinese Academy of Sciences, Chongqing 400714, China (e-mail: wangling1820@126.com).
➢ Y. X. Huang is with the Hanhong College, Southwest University, Chongqing 400715, China (e-mail: hyx200303@sina.com).
➢ H. Wu is with the College of Computer and Information Science, Southwest University, Chongqing 400715, China (e-mail: haowuf@gmail.com).

b) introducing three transformation schemes into the GTCN for aggregating the temporal information; and
c) building the ensemble of diverse transformation schemes to obtain high representation capacity.

This study mainly contributes as follows:
a) We propose a STGCNDT model. It integrates three different transformation schemes, which can capture complex temporal patterns in dynamic graphs and enhance the representation capabilities of dynamic graphs.
b) The GTCN model with different transformation schemes was introduced and investigated, and performance on extensive DG datasets was comparatively analyzed.

## II. PRELIMINARIES

### A. Problem Formulation

In this paper, a DG is defined as $G = \{G^1, G^2, \ldots, G^T\}$ and each graph snapshot $G^t = (V, E^t, X^t)$ is a graph at time slot $t$ ($1 \leq t \leq T$), where $V$ is the node-set whose number of nodes is $N=|V|$. $E^t$ represents the edge set at time slot $t$, and $X^t$ contains all nodes and their feature vectors. Specifically, for a graph snapshot $G^t$, $E^t$ can be represented as an adjacency matrix $A^t$, whose element $a_{ij}^t$ is one if the edge between node $i$ and $j$ exists, and zero otherwise [26-29]. A DG can be described by an adjacency tensor $\mathbf{A}=[a_{ij}^t]$ with the size of $(N \times N \times T)$ to reserve the spatial-temporal pattern. Given a DG, the goal of representation is to obtain a representation tensor $\mathbf{H}$.

### B. Tensor M-Product Framework

The main tensor calculations used in this paper are as follows:

**Mode-n product**. The mode-n product generalizes the matrix-matrix product to the tensor-matrix product. Given a tensor $\mathbf{X} \in \mathbb{R}^{I_1 \times \ldots \times I_{n-1} \times I_n \times \ldots \times I_N}$ and a matrix $U^{D \times I_n}$, then the mode-n product obtains a tensor $(\mathbf{X} \times_n U) \in \mathbb{R}^{I_1 \times \ldots \times I_{n-1} \times D \times \ldots \times I_N}$. Its element at position $(i_1, \ldots, i_{n-1}, d, i_{n+1}, \ldots, i_N)$ is defined as:

$$(\mathbf{X} \times_n U)_{i_1,\ldots,i_{n-1},d,i_{n+1},\ldots,i_N} = \sum_{i_n}^{I_n} U_{d,i_n} X_{i_1,\ldots,i_{n-1},i_n,i_{n+1},\ldots,i_N}. \quad (1)$$

**M-transform**. Given a transform matrix $M \in \mathbb{R}^{T \times T}$. The M-transform of a tensor $\mathbf{X} \in \mathbb{R}^{I \times J \times T}$ is denoted by $\mathbf{X} \times_3 M$ and is defined as:

$$(\mathbf{X} \times_3 M)_{i,j,t} = \sum_{k=1}^{T} M_{t,k} X_{i,j,k}. \quad (2)$$

Note the (2) can be also written in a matrix form, i.e., $(\mathbf{X} \times_3 M) = \text{fold}(M \text{unfold}(\mathbf{X}))$, where the unfold operation takes the tubes of $\mathbf{X}$ and stacks them into matrix columns with the dimension of $T$-$IJ$ as fold (unfold ($\mathbf{X}$))=$\mathbf{X}$. It should be pointed out that $(\mathbf{X} \times_3 M) \times_3 M^{-1}=\mathbf{X}$, if $M$ is an invertible matrix.

**Face-wise Product**. Given two three-order tensors $\mathbf{X} \in \mathbb{R}^{I \times J \times T}$ and $\mathbf{Y} \in \mathbb{R}^{J \times K \times T}$, the face-wise product is denoted by $(\mathbf{X} \otimes \mathbf{Y}) \in \mathbb{R}^{I \times K \times T}$ and is defined as:

$$(\mathbf{X} \otimes \mathbf{Y})_{:,:,t} = \mathbf{X}_{:,:,t} \mathbf{Y}_{:,:,t}. \quad (3)$$

**M-product**. Given two three-order tensors $\mathbf{X} \in \mathbb{R}^{I \times J \times T}$ and $\mathbf{Y} \in \mathbb{R}^{J \times K \times T}$, and an invertible matrix $M \in \mathbb{R}^{T \times T}$, the M-product is denoted by $(\mathbf{X} * \mathbf{Y}) \in \mathbb{R}^{I \times K \times T}$ and is defined as:

$$\mathbf{X} * \mathbf{Y} = \left( (\mathbf{X} \times_3 M) \otimes (\mathbf{Y} \times_3 M) \right) \times_3 M^{-1}. \quad (4)$$

## III. THE PROPOSED STGCNDT MODEL

### A. Graph Tensor Convolutional Network

In this study, we proposed a GTCN model in which the spatiotemporal information of DG is modeled as a whole, avoiding the separation and loss of information as shown in Fig. 2. Specifically, the graph tensor convolution network is designed based on the M-product as:

$$\mathbf{H} = \sigma(\mathbf{A} * \mathbf{X} * \mathbf{W}), \quad (5)$$

where $\sigma$ is a non-linear activation function like Sigmoid, $\mathbf{A}$ is the adjacent tensor, $\mathbf{X}$ is the node feature tensor, $\mathbf{W}^{F \times F \times T}$ is a learnable weight tensor for feature transformation and $\mathbf{H}^{N \times F \times T}$ is the node representation tensor.

The GTCN can aggregate the information from spatio and temporal dimensions of a DG tensor directly since it integrates the tensor face-wise product and the tensor M-transform. The former can implement spatio message passing on each front slice of $\mathbf{X}$ with the help of the adjacent tensor $\mathbf{A}$, and the latter is used to implement temporal message passing through the transform matrix $M$. Specifically, corresponding to GTCN with the tensor form, the spatiotemporal message passing of each node $i$ at time $t$ can be formulated as:

$$\begin{cases} c_i^t = \sum_{j \in N(i) \cup \{i\}} \Phi(a_{ij}^t) \Phi(x_j^t), \\ x_i^t = \sigma(c_i^t W^t), \end{cases} \quad (6)$$

where $c_i^t$ is the message vector from the neighbor set $N(i)$ of node $i$, and $a_{ij}^t$ is the entry of the adjacent tensor **A** at position $(i, j, t)$, and $x_i^t$ is the node feature vector. As the $t$-th front slice of **W**, $W^t$ converts the message vector at $t$-th time slot into node features $x_i^t$. The function $\Phi(\cdot)$ represents M-Transform for aggregating node temporal features as:

$$\Phi(x_j^t) = \sum_{k=1}^{T} m_{tk} x_j^k, \tag{7}$$

where $m_{tk}$ is the value of transform matrix $M$, which gives the weight of $k$-th temporal information aggregated to $t$-th temporal information. $x_j^k$ is the feature vector of node $j$ at time slot $k$.

From the message-passing perspective, it is clear that the matrix $M$ of the transformation scheme is the key to GTCN since it determines the way of temporal message passing. Thus, GTCN with improper temporal transformation schemes may suffer from low presentation ability due to the complex temporal dynamics in DGs. To boost the representation learning, GTCN with three temporal transformation schemes, i.e., Discrete Fourier Transform [26], Discrete cosine transform [27], and Haar Wavelet Transform [28], are proposed and investigated as follows.

*B. GTCN with Discrete Fourier Transform*

The Discrete Fourier Transform (DFT) is a fundamental tool in the field of signal processing and analysis, providing a bridge between temporal information in the time domain and spectral information in the frequency domain. It can effectively handle signals that change over time. Specifically, DFT is defined as:

$$\hat{s}^t = \frac{1}{\sqrt{T}} \sum_{k=1}^{T} \exp\left(-2i\pi t \frac{(k-1)}{T}\right) s^k, \tag{8}$$

where $\hat{x}^t$ represents the information at time $t$ transformed by DFT, $x^k$ is the input information sequence, and i is the imaginary unit. Then, the transform matrix $M_{DCT}=[m_{tk}]^{T \times T}$ of DFT is:

$$m_{tk} = \frac{1}{\sqrt{T}} \exp\left(-2i\pi t \frac{(k-1)}{T}\right). \tag{9}$$

When GTCN adopts DFT as a transformation scheme (named GTCN$_{DFT}$), we obtain the DG representation tensor **H**$_{DFT}$.

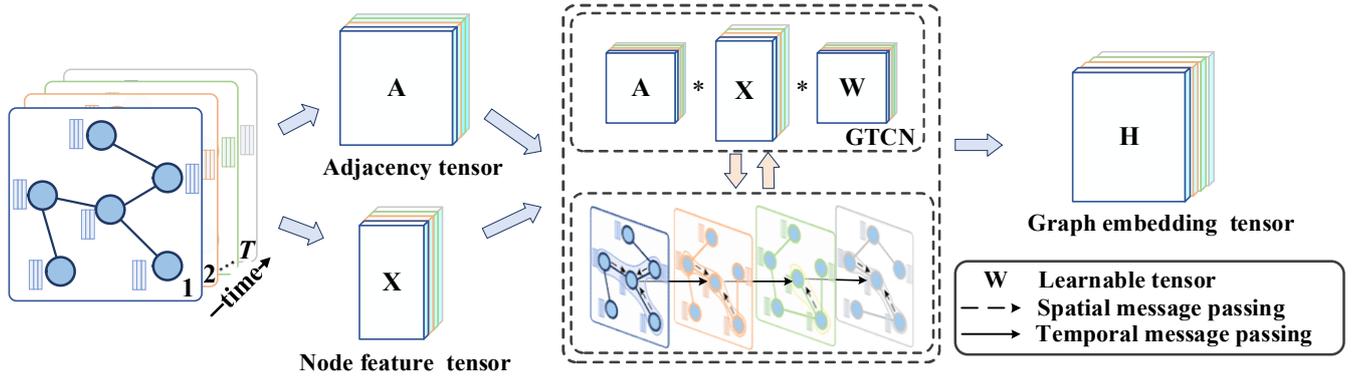

Fig. 2. The proposed GTCN framework.

*C. GTCN with Discrete Cosine Transform*

The Discrete Cosine Transform (DCT) is also a powerful tool in temporal signal processing. It is similar to the DFT, while the DFT uses complex exponentials to represent signals, the DCT uses only cosine functions, which makes it more suitable for real-valued temporal signals. The basic idea behind the DCT is to represent a temporal signal as a weighted sum of cosine functions of different frequencies. By doing so, it separates the signal into different frequency components. DCT is expressed as:

$$\hat{s}^t = \sum_{k=1}^{T} \alpha(k-1) \cos\left[\frac{\pi}{T} t\left((k-1) + \frac{1}{2}\right)\right] s^k, \tag{10}$$

where $\hat{s}^t$ represents the information at time $t$ transformed by DCT, and $s^k$ is the input information sequence. The coefficient $\alpha(n)$ in DCT is defined as:

$$\alpha(n) = \begin{cases} \sqrt{\dfrac{1}{T}} & n = 0, \\ \sqrt{\dfrac{2}{T}} & 1 \leq n \leq T. \end{cases} \tag{11}$$

Then, the transform matrix $M_{DCT}=[m_{tk}]^{T \times T}$ for DCT is set as:

$$m_{tk} = \alpha(k-1)\cos\left[\frac{\pi}{T}t\left((k-1)+\frac{1}{2}\right)\right], \quad (12)$$

By adopting $M_{DCT}$ in the GTCN (named GTCH$_{DCT}$), representation tensor $\mathbf{H}_{DCT}$ of DG is obtained.

*D. GTCN with Haar Wavelet Transform*

The Haar Wavelet Transform was introduced as a signal processing method different from traditional Fourier analysis, which is based on the Haar Wavelet. The HWT decomposes a signal into a set of approximation coefficients and detail coefficients, providing a multi-resolution analysis of the signal. Specifically, HWT is defined as:

$$\hat{s}^t = \sum_{k=1}^{T} \Psi_i^j(k-1)s^k, \quad (13)$$

where $\Psi_i^j(\cdot)$ represents the HWT basis function, which is expressed as:

$$\Psi_i^j(t) = \Psi(2^j t - i), \quad 0 \le j \le T-1, 0 \le i \le 2^j -1, \quad (14)$$

where $j$ is the scale factor, and $i$ is the translation parameter. The above formula shows that $\Psi_i^j(\cdot)$ is obtained from $\Psi(\cdot)$ by scaling and shifting, and $\Psi(\cdot)$ is defined as:

$$\Psi(z) = \begin{cases} 1 & 0 \le z \le 0.5, \\ -1 & 0.5 \le z \le 1, \\ 0 & \text{otherwise.} \end{cases} \quad (15)$$

Thus, the transform matrix $M_{HWT}=[m_{tk}]^{T\times T}$ for HWT is set as:

$$m_{tk} = \Psi_i^j(k-1), \quad (16)$$

By adopting $M_{HWT}$ in the GTCN (named GTCN$_{HWT}$), a representation tensor $\mathbf{H}_{HWT}$ of DG is obtained.

*E. Ensemble Module*

Ensemble learning is a method that integrates multiple learning algorithms to improve representation learning capabilities [29-31]. In this paper, we introduce three different transformation schemes for GTCN, from which three graph representation tensors $\mathbf{H}_{DFT}$, $\mathbf{H}_{DCT}$, and $\mathbf{H}_{HWT}$ are obtained. To improve the model's accuracy, an ensemble model STGCNDT is proposed. Specifically, different ensemble weights are assigned for each representation tensor to obtain ensemble presentation tensor $\mathbf{H}$ as:

$$\mathbf{H} = \alpha \mathbf{H}_{DFT} + \beta \mathbf{H}_{DCT} + \chi \mathbf{H}_{HWT}, \quad (17)$$

where $\alpha$, $\beta$, and $\chi$ are the ensemble weights of different representation tensors. In this study, we set $\alpha = \beta = \chi = 1/3$.

*F. Model Optimization*

This study utilizes the estimation of missing link weights as the downstream task to verify the representation ability of the proposed STGCNDT. After obtaining the ensemble node embedding tensor $\mathbf{H}$, the link weight between node $i$ and $j$ at time slot $t$ is estimated as follows:

$$\hat{y}_{ij}^t = \left[h_i^t \| h_j^t\right] r^\top, \quad (18)$$

where $\hat{y}_{ij}^t$ is the estimation weight, $r$ is a regression learnable vector to aggregate the features as an estimation value, and $[\cdot\|\cdot]$ denotes the concatenation operation of embedding vectors.

To optimize model parameters, a learning objective defined on known data following the data density-oriented modeling principle [32-39] is built as:

$$L(\Theta) = \sum_{y_{ij}^t \in \Lambda} \left(y_{ij}^t - \hat{y}_{ij}^t\right)^2 + \kappa \|\Theta\|_2, \quad (19)$$

where $y_{ij}^t$ is the true value of link weight and $\Theta = \{\mathbf{W}, r\}$ represents the parameter set in the model to optimize. $\Lambda$ is the training set. $\|\cdot\|_2$ is $L_2$ regularization with coefficient value $\kappa$.

IV. EMPIRICAL STUDIES

*A. General Settings*

**Evaluation Protocol.** This study focuses on dynamic graph representation learning and takes the link weight estimation problem as a downstream analysis task. Thus, the estimation accuracy is adopted as the evaluation protocol. Mean Absolute Error (MAE) [40-62] and Root Mean-Squared Error (RMSE) [76-80] are commonly adopted to measure the estimation accuracy of a tested model [38-46], and they are computed as:

$$\text{MAE} = \left(\sum_{y_{ij}^t \in \Omega} \left|y_{ij}^t - \hat{y}_{ij}^t\right|_{abs}\right) \Big/ |\Omega|,$$

$$\text{RMSE} = \sqrt{\left(\sum_{y_{ij}^t \in \Omega} \left(y_{ij}^t - \hat{y}_{ij}^t\right)^2\right) \Big/ |\Omega|},$$

where $|a|_{abs}$ obtains $a$'s absolute value and $\Omega$ indicates the test set whose number of entries is denoted by $|\Omega|$.

**Datasets.** Four DG datasets from personal dynamic terminal commutation networks are adopted in the experiment. The detailed statistical information of each dataset is shown in Table I. Specifically, the four DG datasets are consistently divided into training, validation, and testing sets in a 60%:20%:20% ratio.

TABLE I. EXPERIMENT DATASET DETAILS.

| Dataset | Nodes | Edges | Time slots |
|---|---|---|---|
| D1 | 3,871 | 57,651 | 65 |
| D2 | 6,924 | 120,103 | 73 |
| D3 | 11,036 | 164,823 | 63 |
| D4 | 14,040 | 192,107 | 97 |

**Baselines.** To evaluate the proposed model, we compare it with four baseline graph representation learning models, which include one static graph convolution model (GCN [24]), Three dynamic graph convolution models (WDGCN [26], and EvolveGCN [31], STTGCN [85]).

**Training Settings.** a) Adam optimizer is adopted and the termination conditions of training are consistent for all the compared modelsAn early stop setting is set, that is, when the error on the verification set increases ten times continuously, the training will be stopped. b) The dimension of node embedding is 20 for all involved models.

### B. Comparison Performance

TABLE II. THE COMPARISON RESULTS ON MAE/RMSE.

| No. | Metric | GCN | WDGCN | EvolveGCN | STTGCN | GTCN$_{DFT}$ | GTCN$_{DCT}$ | GTCN$_{HWT}$ | STGCNDT |
|---|---|---|---|---|---|---|---|---|---|
| D1 | MAE | 0.14737$_{\pm1E-2}$ | 0.13513$_{\pm7E-3}$ | 0.13797$_{\pm4E-3}$ | 0.13900$_{\pm8E-3}$ | 0.12027$_{\pm2E-3}$ | 0.12243$_{\pm4E-3}$ | 0.12719$_{\pm2E-3}$ | 0.11513$_{\pm8E-3}$ |
| | RMSE | 0.20453$_{\pm8E-3}$ | 0.19590$_{\pm2E-3}$ | 0.18777$_{\pm6E-3}$ | 0.19253$_{\pm1E-2}$ | 0.17997$_{\pm8E-3}$ | 0.17572$_{\pm6E-3}$ | 0.17623$_{\pm2E-3}$ | 0.17093$_{\pm5E-3}$ |
| D2 | MAE | 0.14427$_{\pm8E-3}$ | 0.13589$_{\pm6E-3}$ | 0.13491$_{\pm4E-3}$ | 0.13442$_{\pm9E-3}$ | 0.12753$_{\pm1E-2}$ | 0.12798$_{\pm1E-3}$ | 0.12755$_{\pm2E-2}$ | 0.11912$_{\pm6E-3}$ |
| | RMSE | 0.21273$_{\pm3E-3}$ | 0.19502$_{\pm5E-3}$ | 0.20725$_{\pm5E-3}$ | 0.21273$_{\pm3E-3}$ | 0.19007$_{\pm6E-3}$ | 0.19053$_{\pm7E-3}$ | 0.20280$_{\pm6E-3}$ | 0.18681$_{\pm5E-3}$ |
| D3 | MAE | 0.13613$_{\pm4E-3}$ | 0.13877$_{\pm2E-3}$ | 0.13545$_{\pm9E-3}$ | 0.13727$_{\pm8E-3}$ | 0.12793$_{\pm7E-3}$ | 0.12776$_{\pm3E-3}$ | 0.12395$_{\pm4E-3}$ | 0.11883$_{\pm2E-3}$ |
| | RMSE | 0.20623$_{\pm1E-3}$ | 0.19502$_{\pm7E-3}$ | 0.18733$_{\pm7E-3}$ | 0.19294$_{\pm6E-3}$ | 0.18287$_{\pm1E-2}$ | 0.17630$_{\pm5E-3}$ | 0.17987$_{\pm7E-3}$ | 0.17257$_{\pm2E-3}$ |
| D4 | MAE | 0.13357$_{\pm6E-3}$ | 0.13049$_{\pm8E-3}$ | 0.13120$_{\pm1E-2}$ | 0.12878$_{\pm7E-3}$ | 0.12377$_{\pm9E-1}$ | 0.11997$_{\pm2E-3}$ | 0.12263$_{\pm6E-3}$ | 0.11677$_{\pm5E-3}$ |
| | RMSE | 0.21450$_{\pm7E-3}$ | 0.20960$_{\pm1E-3}$ | 0.20539$_{\pm1E-2}$ | 0.19732$_{\pm5E-3}$ | 0.18961$_{\pm1E-2}$ | 0.19103$_{\pm2E-3}$ | 0.18530$_{\pm6E-3}$ | 0.17926$_{\pm9E-3}$ |

We compare the performance of all models in this section. Table II show the comparison results on MAE and RMSE. From them, we have the following findings:

a) STGCNDT shows outstanding advantages in the link weight estimation of DG. According to Table II and Fig. 3, the STGCNDT obtains the lowest estimation errors in all testing cases. For instance, on D1, STGCNDT obtains the optimal MAE at 0.11513, which is 21.87% lower than 0.14737 obtained by GCN, 14.80% lower than 0.13513 obtained by WDGCN, 16.55% lower than 0.13797 obtained by EvolveGCN, 17.17% lower than 0.13900 obtained by STTGCN. Similar outcomes are acquired in the other datasets as shown in Table II and Fig. 3. In general, STGCNDT's representation ability to the DG is notable.

b) Compared with the GCN model, STGCNDT not only utilizes the information of spatio neighbors but also captures the temporal dynamic of nodes, which is crucial for DG representation learning. Moreover, STGCNDT demonstrates superior performance compared to the dynamic GCN models, i.e., WDGCN, EvolveGCN, and STTGCN, that is because STGCNDT adopts high-order tensors to uniformly describe the spatio-temporal information of DG, avoiding the loss of information.

### B. Different Transform Schemes Analyses

In this section, we validate the effects of diversified transformation schemes. The results on D1-D4 are shown in Table II and Fig. 3. GTCN$_{DFT}$, GTCN$_{DCT}$, and GTCN$_{HWT}$ correspond to adopting DFT, DCT, and HWT transformation schemes in GTCN, respectively. From the results, we have:

a) STGCNDT demonstrates excellent performance compared with GTCN$_{DFT}$, GTCN$_{DCT}$, and GTCN$_{HWT}$, which verifies that the ensemble method can effectively improve the representation ability. Taking the results on D1 as an example, the MAE of STGCNDT is 0.11513, which is 4.27%, 5.96%, and 9.48% lower than 0.12027, 0.12243, and 0.12719 obtained by GTCN$_{DFT}$, GTCN$_{DCT}$, and GTCN$_{HWT}$, respectively. Moreover, the RMSE of STGCNDT is 0.17093. Compared with GTCN$_{DFT}$, GTCN$_{DCT}$, and GTCN$_{HWT}$, its gain is 5.02%, 2.72%, and 3.01%. Similar outcomes are also funded in the other datasets. It can be seen from the above analysis that aggregating the diversified transformer schemes into the model is beneficial to boost the DG's representation ability.

b) The optimal single transformation scheme is related to the datasets. For example, on MAE, the optimal settings for D1 and D2 are DFT, while the optimal settings for D3 and D4 are HWT and DCT respectively. Specifically, on D1, the MAE of $GTCN_{DFT}$ is 0.12027, which is 1.76% and 5.44% lower than 0.12243 and 0.12719 obtained by $GTCN_{DCT}$ and $GTCN_{HWT}$, respectively. The above findings indicate that different transformation schemes affect the performance of GTCN, and appropriate transformation schemes can improve the representation ability of GTCN.

## V. Conclusions

In this paper, we propose a Spatial-temporal Graph Convolutional Networks with Diversified Transformation (STGCNDT). It is integrated with three different transformation schemes to model the complex temporal patterns of DG. Thus, STGCNDT obtains significant performance improvements on the link weight estimation task in DGs. Moreover, a comparative analysis of different transformation schemes also is conducted, which demonstrates the importance of appropriate transformation schemes for GTCN.

In the future, we plan to further improve the representation ability of STGCNDT by adopting an adaptive ensemble scheme to measure the importance of different transformation schemes. Moreover, we also plan to theoretically analyze why the ensemble model obtains a better representation learning of DG.